\documentclass[conference]{IEEEtran}
\IEEEoverridecommandlockouts
\usepackage{cite}
\usepackage{amsmath,amssymb,amsfonts}
\usepackage{algorithmic}
\usepackage{graphicx}
\usepackage{comment}
\usepackage{tabularx}
\usepackage{textcomp}
\usepackage{rotating}
\usepackage{multirow} 
\usepackage{booktabs}
\usepackage{xcolor, soul}
\usepackage{xspace}
\usepackage[a4paper, total={184mm,239mm}]{geometry}
\DeclareMathOperator*{\argmaxA}{arg\,max} %
\DeclareMathAlphabet{\mathcal}{OMS}{cmsy}{m}{n}
\def\BibTeX{{\rm B\kern-.05em{\sc i\kern-.025em b}\kern-.08em
    T\kern-.1667em\lower.7ex\hbox{E}\kern-.125emX}}
\begin{document}
\newcommand{\modelA}{\emph{HDC-ZSC}\xspace}
\newcommand{\VSA}{\emph{HDC}\xspace}
\newcommand{\modelB}{\emph{Trainable-MLP model}\xspace}
\title{Zero-shot Classification using Hyperdimensional Computing}

\author{\IEEEauthorblockA{Authors removed for the double-blind review}}

\author{\IEEEauthorblockN{Samuele Ruffino\IEEEauthorrefmark{1}\IEEEauthorrefmark{2},
Geethan Karunaratne\IEEEauthorrefmark{1},
Michael Hersche\IEEEauthorrefmark{1}\IEEEauthorrefmark{2}, \\
Luca Benini\IEEEauthorrefmark{2},
Abu Sebastian\IEEEauthorrefmark{1}, and
Abbas Rahimi\IEEEauthorrefmark{1}}\\
\IEEEauthorblockA{\IEEEauthorrefmark{1}IBM Research -- Zurich, Switzerland 
\IEEEauthorrefmark{3}ETH Zurich, Switzerland}
\thanks{
Corresponding author: Geethan Karunaratne (email: kar@zurich.ibm.com)
}
}

\maketitle

\begin{abstract}
Classification based on Zero-shot Learning (ZSL) is the ability of a model to classify inputs into novel classes on which the model has not previously seen any training examples.
Providing an auxiliary descriptor in the form of a set of attributes describing the new classes involved in the ZSL-based classification is one of the favored approaches to solving this challenging task.
In this work, inspired by Hyperdimensional Computing (\VSA{}), we propose the use of stationary binary codebooks of symbol-like distributed representations inside an attribute encoder to compactly represent a computationally simple end-to-end trainable model, which we name Hyperdimensional Computing Zero-shot Classifier~(\modelA).
It consists of a trainable image encoder, an attribute encoder based on HDC, and a similarity kernel.
We show that \modelA can be used to first perform zero-shot attribute extraction tasks and, can later be repurposed for Zero-shot Classification tasks with minimal architectural changes and minimal model retraining.
\modelA achieves Pareto optimal results with a 63.8\% top-1 classification accuracy on the CUB-200 dataset by having only 26.6 million trainable parameters. 
Compared to two other state-of-the-art non-generative approaches, \modelA achieves 4.3\% and 9.9\% better accuracy, while they require more than 1.85$\times$ and 1.72$\times$ parameters compared to \modelA, respectively.

\end{abstract}


\section{Introduction}
Object recognition using classical machine learning approaches heavily depends on the availability of a sufficiently large labeled training dataset.
Particularly for fine-grained classification tasks, creating high-quality annotations is expensive and time-consuming; therefore, models with generalization capability to classify images from unseen (or partially seen) classes have been studied over the past few years. 
Few-shot Learning has recently emerged as a popular topic of research in this regard, which attempts to rapidly train the existing models by admitting only one or very few instances of training data per class~\cite{matchingnets,snell2017prototypical,sunder2019oneshot}.

In an extreme case, Zero-shot Learning (ZSL) aims at classifying an object without having seen any instances from the same class during training~\cite{romera2015ESZSL, Altszyler2020zslmultidomain, norouzi2013zero, DBLP:journals/corr/KodirovXG17, DBLP:journals/corr/ZhangS15d, huynh2021composer, DBLP:journals/corr/abs-2003-07833}. 
To enable the recognition of previously unseen classes, in ZSL, the learner is provided with a unique descriptor that distinguishes the unseen class with respect to the classes the learner was trained on. 
The descriptor is typically a collection of attributes, each of which can take a finite range of values. Attributes generated per instance can be human-annotated or the captions accompanying the images~\cite{radford2021clip}. 
A class-level descriptor can be generated by observing a sample statistic from all instance-level attribute value distributions of the class concerned.
The goal of ZSL is, having seen certain combinations of values for the set of attributes, to successfully interpolate to other value combinations corresponding to the unseen classes and thereby inferring the classes of the input examples provided.

\begin{figure}
  \centering
  \includegraphics[width=1\linewidth]{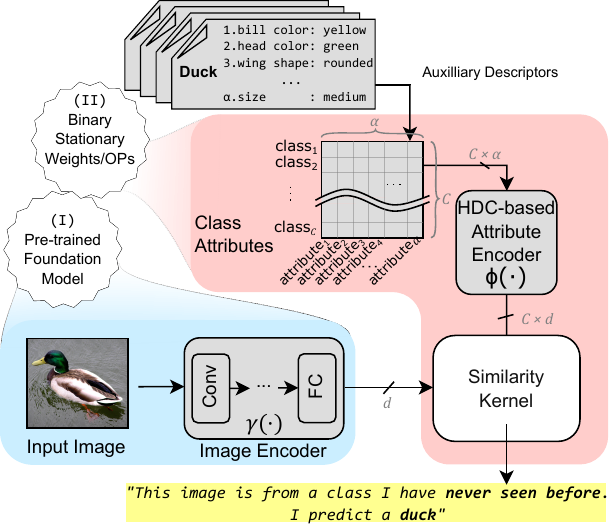}
  \caption{The general model structure employed for Zero-shot Classification in this work. It comprises two main modules: a pre-trained foundation model image encoder and an HDC-based attribute encoder. The attribute encoder consists of weights with fixed binary vectors and binary vector operations, providing opportunities for implementation in resource-constrained edge devices. Modules filled with gray color remain stationary during inference.}
   \label{fig:VSA_class_pred_general}
\end{figure}

A ZSC model generally consists of an image encoder and an attribute encoder as shown in Fig.~\ref{fig:VSA_class_pred_general}.
At inference time, the image encoder accepts an image from an \emph{unseen} class on which the model was not trained. In this example, a duck image is presented to the model, but no images from the duck class were provided during training. 
Therefore it happens to be an input from a class that the model has neither seen nor tuned to before.
However, the model gets access to auxiliary information of all classes including the unseen ones via the attribute encoder.
In this example, the attributes unique to the class of duck such as bill color, head color, wing shape, size, etc. are fed in as an auxiliary descriptor for the duck class. 
Having seen the attributes in different combinations for other classes during training, a ZSL model infers that the attributes present in the given image belong to the description of the duck class.

There are several approaches for solving ZSL problems, which can be broadly divided into two categories: non-generative and generative. In non-generative approaches, mapping functions are explored for image and attribute descriptions such that their alignments can be generalized by applying batch-level contrastive loss functions~\cite{radford2021clip,romera2015ESZSL}. The generative approaches on the other hand rely on larger models capable of artificially manufacturing instances for a given unseen class descriptor thereby helping effectively turn the problem into a few-shot learning problem~\cite{8953480}.

This work presents a new non-generative end-to-end training method for Zero-shot Classification (ZSC) featuring attribute encoders containing fixed and compact dictionaries.
Thanks to the theory of Hyperdimensional Computing (\VSA{})~\cite{Kanerva2009}, these compact codebooks, representing values and attribute groups, can be initialized randomly with binary codevectors when sufficiently high dimensionality is given.
\VSA has recently been combined with neural networks which not only achieved the state-of-the-art accuracy in image-based few-shot learning~\cite{kar_ncom_2021} and few-shot continual learning~\cite{C-FSCIL_CVPR22,kar2022fscl} but also led to energy-efficient hardware implementations. Here, we further expand this combination to the ZSC tasks.

The contributions of this work are listed below:
\begin{itemize}
    \item Introducing a novel hybrid architecture for ZSC named \modelA. It comprises a trainable foundation model image encoder that extracts features/embeddings from images, an \VSA-based fixed (stationary) attribute encoder that transforms auxiliary attributes into high-dimensional binary/bipolar vector embeddings, and finally, a kernel that measures the similarity between the two embeddings, as shown in Fig.~\ref{fig:VSA_class_pred_general}.
    \item Proposing a training methodology to boost ZSC accuracy beyond previous state-of-the-art non-generative methods. It consists of pre-training the model on a standard image classification task at first, followed by a domain-specific attribute extraction task. Here we predict the attributes present in an image to match the ground-truth attributes of the target.
    In the final phase of training, the matured model is exposed to the ZSC task by retraining to discriminate attribute vectors representing the classes.
    \item Evaluating our model on the CUB-200 dataset on the attribute extraction task first, followed by the ZSC.  
    For ZSC, the final \modelA achieves an average top-1 classification accuracy of 63.8\% on the CUB-200 dataset. 
    Compared to two other non-generative state-of-the-art approaches~\cite{romera2015ESZSL,jian2021tcn}, this is a 4.3\% and 9.9\% improvement in accuracy, yet our model is made up of 1.85$\times$ and 1.72$\times$ fewer parameters compared to the same reference models, respectively.
\end{itemize}

To put these results into perspective, more computationally intensive prominent generative approaches require 1.75$\times$--2.58$\times$ more parameters while their accuracies reach only a meager 3.9\% increase compared to \modelA.
When compared with both state-of-the-art generative and non-generative approaches, \modelA, and its variants, are in the Pareto front with respect to accuracy and model parameter count (see Fig.~\ref{fig:VSA_zsc_results}). 
This could potentially pave the way for its implementation as an application target in low-power embedded platforms~\cite{pulpissimo}.

\section{Background}

\paragraph{Zero-shot Learning}
The human brain is able to learn new concepts by encountering a few examples or applying deductive, inductive, or abductive reasoning. 
However, the training of classical machine learning models, deep learning models, in particular, rely on a large dataset with an adequate number of training examples for each class.
The idea of ZSL stems from the quest for general artificial intelligence models capable of recognizing classes having seen zero or a few instances thereof similar to the way the brain functions.
In fact, some argue that one-shot learning is a specific case of ZSL called \emph{transductive ZSL}, in which the attribute descriptors are provided in an unstructured pictorial form~\cite{Xian2017zslgood}.
In ZSL the learner is first exposed to a training dataset $\mathcal{D}_{r}$ composing of:
(i) the input object set $\mathcal{X}_{r} = \{\mathbf{x}_{1_r},\mathbf{x}_{2_r},...,\mathbf{x}_{N_r}\}$ which are three-dimensional matrices in the case of images, and $N_r$ stands for the number of training samples,
(ii) the set of class labels $\mathcal{Y}_{r}$ associated with the input objects,
(iii) the class attribute matrix $\mathbf{A}_r \in \mathbb{R}^{C_r\,\times\,\alpha}$, which consists of an attribute vector of length $\alpha$, per unique class. $C_r$ represents the total number of unique classes in the training dataset and, $\alpha$ refers to the number of attributes that are identifiable across training and testing datasets.

Similarly, a testing dataset $\mathcal{D}_{e}$ can be defined with $\mathcal{X}_{e}$, $N_e$, $\mathcal{Y}_{e}$, $\mathbf{A}_e$, $C_e$ referring to the testing dataset's input object set, number of samples, class label set, class attribute matrix, and the number of unique classes, respectively.
In ZSL, the set of class labels in the training and testing datasets are mutually exclusive ($\mathcal{Y}_{r} \bigcap \mathcal{Y}_{e} = \emptyset$).

All ZSL methods associate seen and unseen classes through some form of \emph{auxiliary descriptors} which encodes distinguishing properties of objects (e.g., language description, attributes, etc.), and establish links between previously seen data and new unseen data.
Different ZSL methods have been developed in the past years.
The notable types include 
(i)~\emph{Learning Linear Compatibility}
\cite{DBLP:journals/corr/AkataPHS15},  
\cite{romera2015ESZSL}, \cite{DBLP:journals/corr/KodirovXG17},
in which a linear connection is learned between image and attribute encodings,
(ii)~\emph{Learning Non-Linear Compatibility} \cite{DBLP:journals/corr/XianA0N0S16}, \cite{https://doi.org/10.48550/arxiv.1301.3666}, in which the learned connection is non-linear,
(iii)~\emph{Learning Intermediate Attribute Classifiers} \cite{Lampert2014AttributeBasedCF}, in which an attribute predictor's scores are used for class prediction,
(iv)~\emph{Hybrid Models} \cite{norouzi2013zero}, \cite{DBLP:journals/corr/ZhangS15d},  \cite{DBLP:journals/corr/ChangpinyoCGS16}, \cite{DBLP:journals/corr/VermaR17}, which measures semantic similarity between images and attributes.
Additionally, there exist approaches based on Generative Adversarial Network~\cite{DBLP:journals/corr/abs-1903-10132}, \cite{8953480}, \cite{DBLP:journals/corr/abs-1808-00136}, \cite{DBLP:journals/corr/abs-1712-00981}, which employ supplementary networks to generate unseen class examples during training.

This work follows the linear compatibility approach similar to \emph{ESZSL} \cite{romera2015ESZSL} and \emph{CLIP} \cite{radford2021clip}. 
ESZSL uses the squared loss to learn the bilinear compatibility and regularizes the objective w.r.t Frobenius norm, whereas CLIP tries to maximize the cosine similarity of correct image-attribute pairs while minimizing the same for incorrect pairs.

\paragraph{Hyperdimensional Computing (\VSA)}
 \VSA is a vector space model that encodes symbols starting from randomly initialized high-dimensional vectors called \emph{atomic hypervectors}. 
It has been shown that as the dimensionality of the vector space grows, randomly initialized vectors tend to become \emph{quasi-orthogonal} to each other~\cite{Kanerva2009}.
As one use-case of \VSA, the quasi-orthogonal atomic hypervectors representing a key vector can be combined with the corresponding value vector using a \emph{variable binding} operation ($\odot$) to generate a hypervector that represents the compound symbol within the same dimensional space. 
In addition to the binding operation, \VSA is defined by several other elementary algebraic operations such as bundling ($+$), permutation ($\rho$), and unbinding ($\oslash$).
Hardware prototype of HDC showed substantial energy savings and extremely robust operation~\cite{HDC_ISLPED16}.

\section{Methods}

In this work, an end-to-end approach for ZSL is proposed, exploiting class-attribute association and \VSA.
The concept is illustrated in Fig.~\ref{fig:VSA_class_pred_general}. The inputs to the model are provided in two modalities: (i) a batch of images $\mathbf{X} = (\mathbf{x}_{1},\mathbf{x}_{2},...,\mathbf{x}_{B})$ in matrix form with batch size $B$, and (ii) a class attributes matrix $\mathbf{A}=(\mathbf{a}_1,...,\mathbf{a}_C)\in\mathbb{R}^{C\times\alpha}$, where each row is an attribute vector of $\alpha$ dimension with $C$ vectors in total representing all classes. 
The images of dimensions $h,w,l$ are passed through an image encoder $\gamma(\cdot):\mathbb{R}^{h\times w\times l}\rightarrow\mathbb{R}^{d}$ and the attributes are passed through an attribute encoder $\phi(\cdot):\mathbb{R}^{\alpha}\rightarrow\mathbb{R}^{d}$, both of which generate embeddings of the same dimensionality ($d$).
The embeddings are then compared via a bi-similarity kernel \emph{cossim}, that relates image and class embeddings \cite{romera2015ESZSL}, \cite{radford2021clip}, as follows:
\begin{displaymath}\label{eq:CLIP_cossim}
    cossim(\gamma(\mathbf{X}),\phi(\mathbf{A})) = \frac{1}{K} \, \frac{\gamma(\mathbf{X})^T \cdot \phi(\mathbf{A})}{||\gamma(\mathbf{X})|| \,\, ||\phi(\mathbf{A})||},
\end{displaymath} 
where $K$ is the learnable temperature scaling parameter. The resulting similarity matrix of dimension $B\times C$ is used for error computation with respect to the batch of ground-truth labels and updating the image encoder's weights over a series of epochs.
Finally, the predicted class for every unseen queried image $\mathbf{x}$ is computed as
\begin{displaymath}\label{eq:CLIP_cossim2}
   \hat{\mathbf{y}}=\argmaxA_{i\in  \{1,...,C\}} \,  cossim(\gamma(\mathbf{x}),\phi(\mathbf{a}_i)).
\end{displaymath} 

In summary, our approach employs three main computational modules:
\begin{itemize}
\item Image Encoder $\gamma(\cdot)$: extracts embeddings from images;
\item Attribute Encoder $\phi(\cdot)$: extract embeddings from class attributes;
\item Similarity Kernel: measure the similarity (e.g., dot-product, cosine similarity) between image and attribute embeddings. 
\end{itemize}

\subsection{\VSA-based Attribute Extraction} \label{sec:attribute_pred}

\begin{figure}
  \centering
  \includegraphics[width=\linewidth]{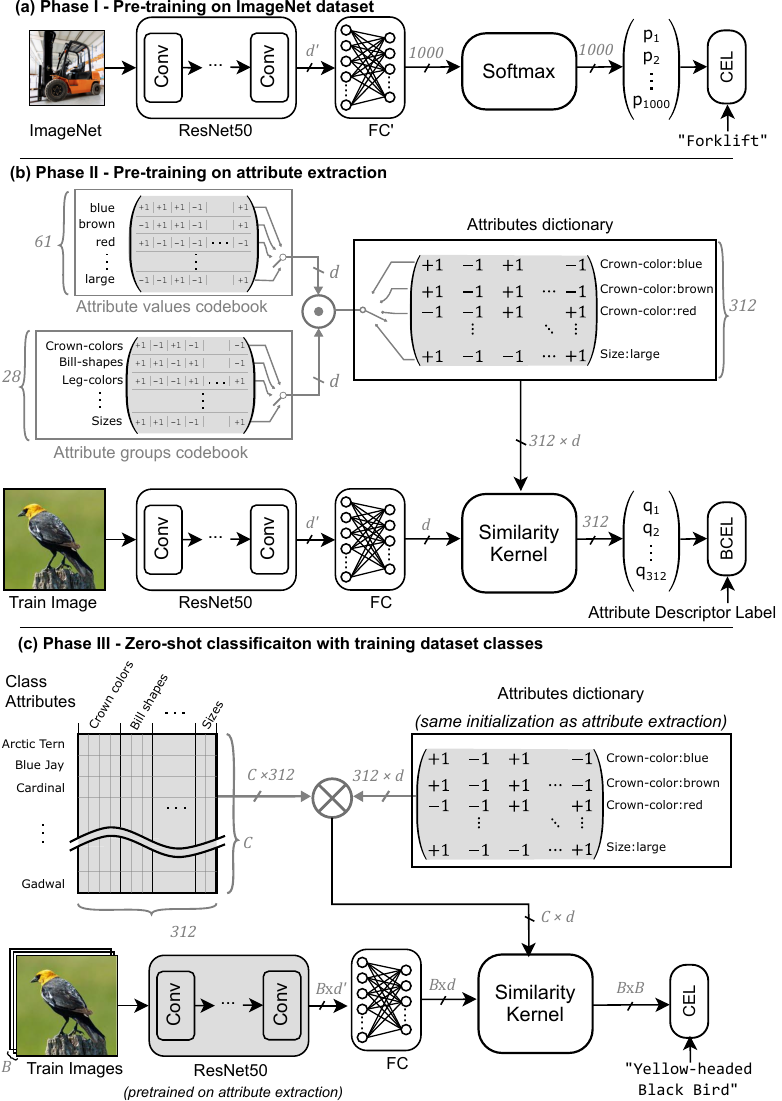}
  \caption{Different phases of Zero-shot Classification training. (a) The backbone network is pre-trained on ImageNet data (b) The backbone network and the projection FC are pre-trained on attribute extraction (c) The backbone network and the projection FC are further fine-tuned with training on Zero-shot Classification task with images of classes in the training dataset. Modules filled with gray color are stationary.}
   \label{fig:VSA_attr_pred}
\end{figure}

The attribute encoder first assigns randomly initialized $d$-dimensional binary/bipolar dense random hypervectors to all atomic level symbols. 
Dense random hypervector generation can be easily realized by sampling from the Rademacher distribution. 
With a sufficiently high dimension, the atomic vectors tend to be quasi-orthogonal to each other.

In this task, we identify atomic hypervectors from two codebooks. They are the \emph{attribute groups} codebook ($\mathbf{g}_1,\mathbf{g}_2,...,\mathbf{g}_G$) and the \emph{attribute values} codebook ($\mathbf{v}_1,\mathbf{v}_2,...,\mathbf{v}_V$). 
By assigning an atomic hypervector for the two sources separately instead of assigning one for each appropriate attribute group/value combination ($\mathbf{b}_1,\mathbf{b}_2,...,\mathbf{b}_\alpha$), we reduce the memory requirement
For example, in the CUB-200 dataset, there are $\alpha=312$ attribute group/value combinations; however, there are only $G=28$ groups (e.g., crown colors, leg colors, sizes, etc.) and $V=61$ unique values (e.g., blue, brown, large, etc.).
This leads to a 71\% reduction in memory requirement for storing atomic hypervectors. 
This results in just 17\,KB of memory for storing the atomic hypervectors which is a negligible amount compared to the image encoder memory requirement which is typically several hundreds of MB.

The next step involves binding appropriate group and value hypervectors together using \emph{variable binding} to materialize the attribute level codevectors on the fly.  
As described in~\cite{hd_optimize}, for dense bipolar hypervectors, this is simply an elementwise multiplication operation (for binary hypervectors the operation becomes elementwise XOR operation) ($\odot$) as in $\mathbf{b}_x = \mathbf{g}_y \odot \mathbf{v}_z $. This shows how $x^{th}\in \{1,...,\alpha\}$ attribute vector $\mathbf{b}_x$ in the attribute dictionary is generated implicitly by binding appropriate $y^{th}\in\{1,...,G\}$ \emph{group} and $z^{th}\in\{1,...,V\}$ \emph{value} hypervectors from the corresponding codebooks.
According to the theory of \VSA, the binding operation produces vectors quasi-orthogonal to the operand vectors. Hence it allows for preserving quasi-orthogonality at the attribute level.

During the attribute prediction, the matrix containing the attribute vectors $\mathbf{B}=~(\mathbf{b}_1,\mathbf{b}_2,...,\mathbf{b}_\alpha)\in\{-1,+1\}^{\alpha\times d}$ is fed to the similarity kernel which additionally
takes the output embedding by the image encoder $\gamma(\mathbf{x})\in\mathbb{R}^{d}$ and
performs cosine similarity calculation as $\mathbf{q} = cossim(\gamma(\mathbf{x}),\mathbf{B})$.

For the image encoder, we use ResNet50 as the backbone network followed by a single fully connected (FC) layer as the image encoder. As shown in Fig.~\ref{fig:VSA_attr_pred}(a) the backbone network is pre-trained with ImageNet1K data by connecting the backbone network to the 1000-dimensional softmax layer via a temporary $FC'$ layer in phase I.
Once this pre-training stage is complete the backbone network is retained with the updated weights for phase II while the $FC'$ layer is replaced with a $FC$ layer which is used to project the backbone network embeddings to a dimension ($d$) preferred by ZSC tasks. The process is pictorially illustrated in Fig.~\ref{fig:VSA_attr_pred}(b). The image encoder after pre-training for the attribute extraction task has matured weights in both the backbone network and the $FC$ layer so that it can be used as the starting point for ZSC in phase III.

There is a large class imbalance in the attribute extraction task due to the dominating number of inactive attributes (e.g., only one color attribute per color group is typically activated at a time among all 15 possible values). 
For this reason, a weighted Binary Cross Entropy Loss (BCEL) is employed between the similarity vector $\mathbf{q}$ and the ground-truth attributes.
The loss is used to update the weights of both the $FC$ layer and the ResNet50 using backpropagation while keeping the attribute encoder designed with codevectors fixed.

\subsection{Zero-shot Classification (ZSC)}
\begin{figure}
  \centering
  \includegraphics[width=\linewidth]{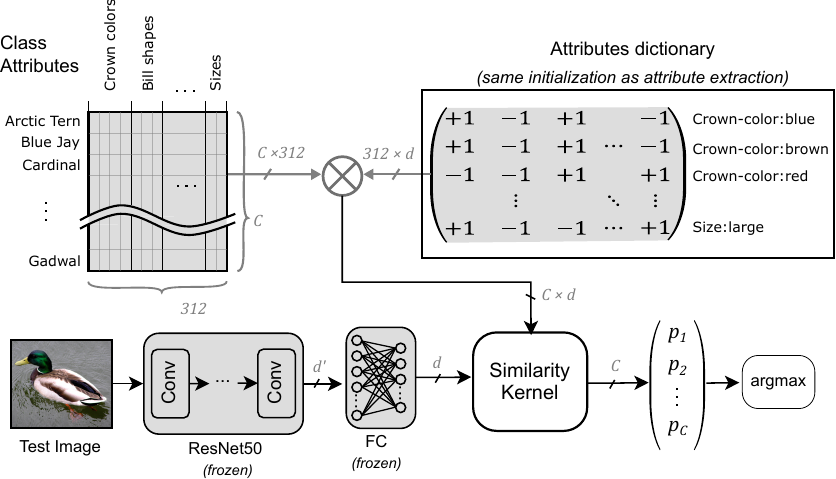}
  \caption{\modelA for Inference in Zero-shot Classification. Weights in both image and attribute encoders are stationary.}
   \label{fig:VSA_zsclass}
\end{figure}

Fig.~\ref{fig:VSA_attr_pred}(c) shows the training procedure of \modelA for ZSC. The image encoder $\gamma(\cdot)$ is pre-trained on ImageNet1K data as well as on the attribute extraction task (see Section~\ref{sec:attribute_pred}).
As the first step, the attribute encoder $\phi(\cdot)$ is defined as the product between the class attributes matrix $\mathbf{A}\in\mathbb{R}^{C\times\alpha}$ and the attribute vector matrix ($\mathbf{B}\in\{-1,+1\}^{\alpha\times d}$), that was populated during attribute extraction task as given in $\phi=\mathbf{A}\times\mathbf{B}$.

The result of the attribute and image encoders are fed to the similarity kernel, which computes class logits via $\mathbf{p} = cossim(\gamma(\mathbf{x}),\phi)$. Cross entropy loss (CEL) between $p$ and the ground-truth class is used to update the $FC$ layer weights in $\gamma$ over several epochs so that the image embeddings are aligned with the attribute embeddings. Having been pre-trained on ImageNet1K and attribute extraction, the backbone network remains stationary during this fine-tuned retraining step.

Finally, the trained model is deployed for ZSC inference as shown in Fig.~\ref{fig:VSA_zsclass}. All weights of the model remain stationary at this stage.

\section{Experiments}

\subsection{Experimental Setup}

\paragraph{Dataset}
We demonstrate the effectiveness of our framework on Caltech-UCSD Birds-200-2011 (CUB-200-2011) \cite{WahCUB_200_2011}, a fine-grained dataset of 200 bird species containing 11,788 images in total. 
The dataset contains continuous class attributes (312-dimensional visual attributes) which are used as the class attribute matrix $\mathbf{A}$. We evaluate our model on two commonly used train-test splits, available in the literature, namely: (i) \emph{noZS} where samples of 100 classes are available in each of train and test datasets, (ii) \emph{ZS} where 150 classes are available for training and the remaining 50 classes for testing.

\paragraph{Evaluation Metrics} 
We use the top-1\% accuracy as a metric for the attribute extraction task. To address the class imbalance issue, we use the \emph{Weighted Mean Average Precision} (WMAP) as another metric. 
WMAP is a modified version of Average Precision designed to compensate for attributes that are less frequent in the dataset.
For ZSC, a commonly used top-1 and top-5 accuracy is employed.

\paragraph{Implementation Details}
We resize input images to 256$\times$256 and augment the training dataset with random rotation in the range of $[-45 ^{\circ}, +45 ^{\circ}]$, center cropping, and random horizontal flip
, before feeding them to the image encoder for both attribute and class extraction task. We implemented our framework in Pytorch and optimized it using \emph{AdamW} optimizer \cite{https://doi.org/10.48550/arxiv.1711.05101} with default settings and \emph{cosine annealing learning rate scheduler} \cite{https://doi.org/10.48550/arxiv.1608.03983}. 
Finally, the results are obtained on the aforementioned metrics by running five trials with different seeds, and average and standard deviation results are reported as: $\mu \pm \sigma$.

\subsection{Experimental Results}

\paragraph{Attribute extraction} Table~\ref{tab:attribute_ext}  summarizes the results of weighted mean average precision (WMAP) and top-1\% accuracy, comparing our approach with Finetag \cite{finetag} and A3M \cite{DBLP:journals/corr/abs-1901-00392} respectively. To facilitate a fair comparison, \emph{noZS} split is used as was the case in the works compared against.
The results show that our \modelA outperforms the previous approaches across both the metrics, by $+4.14\%$ and $+36.71\%$ respectively.

\begin{table}[]
\centering
\caption{Attribute extraction task results comparison with Finetag and A3M using W\_MAP and top-1\% accuracy as evaluation metric}
\label{tab:attribute_ext}
\begin{tabular}{l|rr|rr}
\hline
\multicolumn{1}{c|}{\begin{tabular}[c]{@{}c@{}}Attribute\\Group\end{tabular}} & \multicolumn{1}{c}{\begin{tabular}[c]{@{}c@{}}Finetag\\ (W\_MAP)\end{tabular}} & \multicolumn{1}{c|}{\begin{tabular}[c]{@{}c@{}}Ours\\ (W\_MAP)\end{tabular}} & \multicolumn{1}{c}{\begin{tabular}[c]{@{}c@{}}A3M\\ (top-1\% acc)\end{tabular}} & \multicolumn{1}{c}{\begin{tabular}[c]{@{}c@{}}Ours\\ (top-1\% acc)\end{tabular}} \\ \hline
bill shape                    & 54                                                                             & \textbf{58}                                                                 & 60                                                                              & \textbf{90}                                                                      \\
wing color                    & 57                                                                             & \textbf{60}                                                                 & 45                                                                              & \textbf{90}                                                                      \\
upperpart color               & 55                                                                             & \textbf{57}                                                                 & 43                                                                              & \textbf{90}                                                                      \\
underpart color               & 59                                                                             & \textbf{62}                                                                 & 58                                                                              & \textbf{93}                                                                      \\
breast pattern                & 15                                                                             & \textbf{61}                                                                 & 58                                                                              & \textbf{81}                                                                      \\
back color                    & 50                                                                             & \textbf{53}                                                                 & 45                                                                              & \textbf{91}                                                                      \\
tail shape                    & \textbf{25}                                                                    & \textbf{25}                                                                 & 34                                                                              & \textbf{84}                                                                      \\
uppertail color               & 40                                                                             & \textbf{42}                                                                 & 43                                                                              & \textbf{93}                                                                      \\
head pattern                  & 30                                                                             & \textbf{33}                                                                 & 35                                                                              & \textbf{89}                                                                      \\
breast color                  & 58                                                                             & \textbf{61}                                                                 & 57                                                                              & \textbf{92}                                                                      \\
throat color                  & 57                                                                             & \textbf{61}                                                                 & 60                                                                              & \textbf{93}                                                                      \\
eye color                     & \textbf{76}                                                                    & \textbf{76}                                                                 & 81                                                                              & \textbf{98}                                                                      \\
bill length                   & 73                                                                             & \textbf{76}                                                                 & 72                                                                              & \textbf{80}                                                                      \\
forehead color                & 56                                                                             & \textbf{59}                                                                 & 51                                                                              & \textbf{92}                                                                      \\
tail collor                   & 42                                                                             & \textbf{44}                                                                 & 38                                                                              & \textbf{90}                                                                      \\
nape color                    & 55                                                                             & \textbf{58}                                                                 & 49                                                                              & \textbf{92}                                                                      \\
belly color                   & 58                                                                             & \textbf{61}                                                                 & 59                                                                              & \textbf{93}                                                                      \\
wing shape                    & 24                                                                             & \textbf{25}                                                                 & 32                                                                              & \textbf{80}                                                                      \\
size                          & 55                                                                             & \textbf{56}                                                                 & 58                                                                              & \textbf{81}                                                                      \\
shape                         & 47                                                                             & \textbf{49}                                                                 & 57                                                                              & \textbf{94}                                                                      \\
back pattern                  & 44                                                                             & \textbf{45}                                                                 & 46                                                                              & \textbf{77}                                                                      \\
tail pattern                  & 41                                                                             & \textbf{43}                                                                 & 43                                                                              & \textbf{77}                                                                      \\
belly pattern                 & 60                                                                             & \textbf{62}                                                                 & 62                                                                              & \textbf{81}                                                                      \\
primary color                 & 62                                                                             & \textbf{66}                                                                 & 51                                                                              & \textbf{90}                                                                      \\
leg color                     & 32                                                                             & \textbf{37}                                                                 & 46                                                                              & \textbf{92}                                                                      \\
bill color                    & 42                                                                             & \textbf{47}                                                                 & 47                                                                              & \textbf{91}                                                                      \\
crown color                   & 56                                                                             & \textbf{60}                                                                 & 53                                                                              & \textbf{93}                                                                      \\
wing pattern                  & 48                                                                             & \textbf{50}                                                                 & 48                                                                              & \textbf{72}                                                                      \\ \hline
\textbf{average}              & 48.96                                                                          & \textbf{53.11}                                                              & 51.11                                                                           & \textbf{87.82}   \\ \hline                                                               
\end{tabular}
\end{table}

\begin{figure}[t]
  \centering
  \includegraphics[width=1\linewidth]{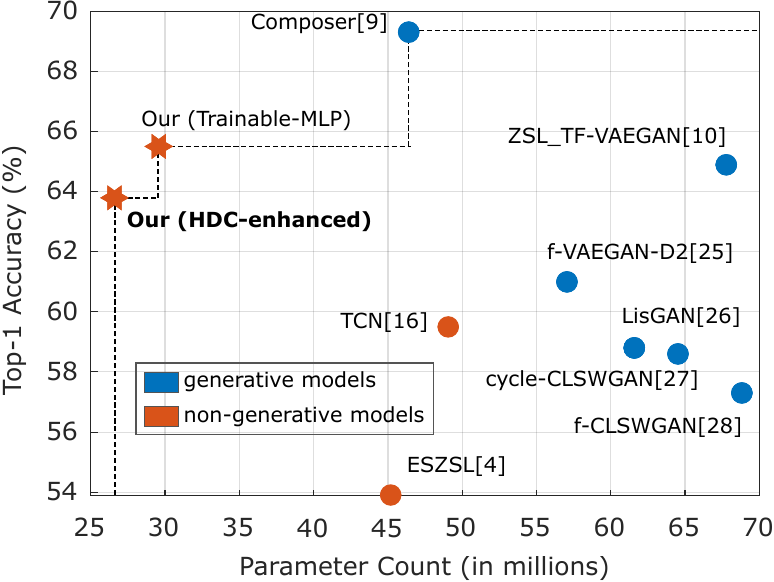}
  \caption{Comparison of Zero-shot Classification accuracy vs model parameter count. Our models \modelA and \modelB are both in the Pareto front.}
   \label{fig:VSA_zsc_results}
\end{figure}

\paragraph{Zero-shot Classification} 
Fig.~\ref{fig:VSA_zsc_results} compares \modelA accuracy performance and number of parameters with other state-of-the-art approaches, both generative and non-generative methods. 
Additionally, results from an alternative \modelB, in which fixed \VSA codebooks-based attribute encoders are replaced by 2-layer trainable MLP, are included in the table for reference.
When compared to generative methods, our models reduce the number of total network parameters by more than 1.75$\times$--2.58$\times$ while maintaining a competitive top-1\% accuracy. Specifically, comparing \modelA with ESZSL~\cite{romera2015ESZSL}, as a non-generative model similar to our approach, we obtain a $+9.9\%$ accuracy improvement while improving the parameter efficiency by more than 1.72$\times$.
As it is shown later in Section~\ref{sec:ablation}, our models train with a minimal number of epochs which enlarges computational cost savings even further compared to the above parameter savings.

\subsection{Ablation study and hyperparameter tuning}\label{sec:ablation}
We present results from training image and attribute encoders of different complexity and maturity in Table~\ref{tab:model_eval}. This result is obtained with a common hyperparameter set without model-wise tuning. From this, we choose the best performing $ResNet50$ with an $FC$ projection layer of $d=1536$ as the preferred configuration of the image encoder for our models. Note that this preferred configuration outperforms the relatively larger $ ResNet101$-based image encoder in terms of Top-1\% accuracy. Further investigations could be carried out as a part of future work to find configurations that suit even more lightweight image encoders.

\begin{table}[]
\centering
\caption{Training of different image and attribute encoders. Pre-training stage II is skipped when the projection $FC$ layer is absent. The figure of merit is Top-1\% accuracy}
\label{tab:model_eval}
\begin{tabular}{ccrrr}
\hline
\multirow{2}{*}{\begin{tabular}[c]{@{}c@{}}Image\\ Encoder\end{tabular}} & \multirow{2}{*}{Pre-train} & \multicolumn{1}{c}{\multirow{2}{*}{$d$}} & \multicolumn{2}{c}{\begin{tabular}[c]{@{}c@{}}Attribute\\ Encoder\end{tabular}}                                                                          \\ \cline{4-5} 
                                                                         &                            & \multicolumn{1}{c}{}                     & \multicolumn{1}{c}{\begin{tabular}[c]{@{}c@{}}HDC\\ ZSC\end{tabular}} & \multicolumn{1}{c}{\begin{tabular}[c]{@{}c@{}}MLP\\ Trainable\end{tabular}} \\ \hline
ResNet50                                                                 & I,III                      & 2048                                     & 55                                                                         & 60                                                                          \\
\multirow{2}{*}{ResNet50+FC}                                             & I,II,III                   & 1536                                     & \textbf{58}                                                                & \textbf{61}                                                                 \\
                                                                         & I,II,III                   & 2048                                     & 50                                                                         & 57                                                                          \\
ResNet101                                                                & I,III                      & 2048                                     & 53                                                                         & 56                                            \\ \hline                             
\end{tabular}
\end{table}

\begin{figure}[t]
  \centering
  \includegraphics[width=1\linewidth]{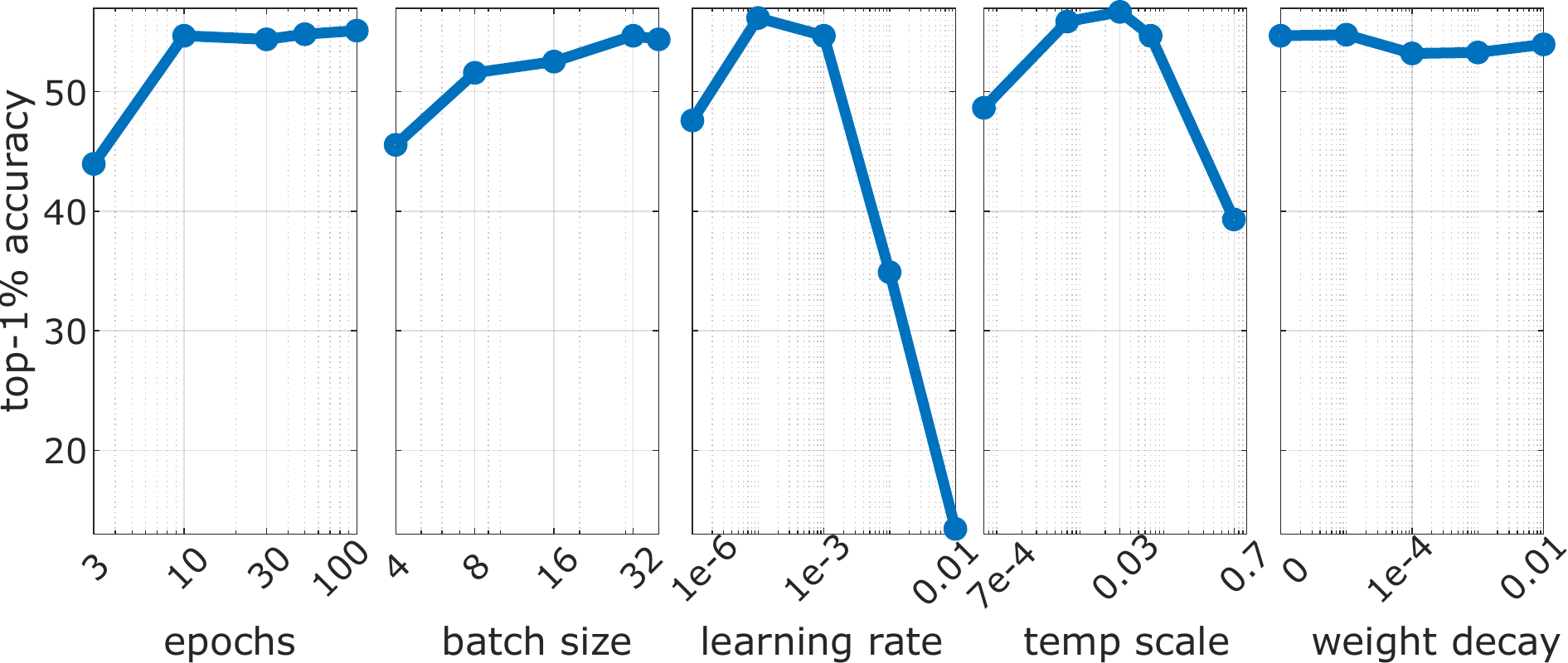}
  \caption{Hyperparameter tuning for \modelA on validation split (50 disjoint classes).}
   \label{fig:hyp-mod1}
\end{figure}

Fig.~\ref{fig:hyp-mod1} provides results from the hyper-parameter exploration for \modelA. The hyperparameters explored include number of epochs, batch size, learning rate, temperature scale, and weight decay. One interesting observation is that the number of epochs required to achieve the best accuracy is around 10 which enables ZSC training to be realized in a reasonable time in low-power hardware systems. The final accuracies are reported with the best-performing hyper-parameter combination is adopted.

\section{Conclusions and outlook}
In this work, we present the concept of integrating lightweight \VSA-enhanced attribute encoders in the backend of non-generative Zero-shot Learning models. 
This approach allows for reducing the model parameter count by representing attributes in compact and stationary codebooks leading to over 1.72$\times$ model parameter storage efficiency while improving top-1 classification accuracy by over 4.3\% compared to other non-generative approaches. 

This outcome opens avenues to explore hardware implementations of the model for both inference and minimal training in low-power heterogeneous embedded systems such as the one described in~\cite{garofalo2022ima}.
For example, the convolutional image encoder can be processed in a general-purpose processor cluster, or any convolutional accelerator, whereas the stationary binary attribute encoder weights and similarity kernel operations can be offloaded to analog~\cite{khaddam2022hermes} or digital~\cite{eggimann20215} non-von Neumann accelerators.

\section*{Acknowledgments}
We would like to acknowledge the early technical investigations by Junyuan Cui.

\bibliographystyle{IEEEtran}
\bibliography{reference}

\end{document}